\documentclass[10pt,twocolumn,letterpaper]{article}

\usepackage{cvpr}              % To produce the CAMERA-READY version
\usepackage{cuted}
\usepackage{algorithm}
\usepackage{algorithmic}
\usepackage{mathrsfs}
\usepackage{adjustbox}
\usepackage[table]{xcolor}
\usepackage{multirow} 
\usepackage{tabularx}

\definecolor{cvprblue}{rgb}{0.21,0.49,0.74}
\usepackage[pagebackref,breaklinks,colorlinks,allcolors=cvprblue]{hyperref}

\newcommand{\VideoMem}{\texttt{VideoMem}\xspace}

\def\paperID{****} % *** Enter the Paper ID here
\def\confName{CVPR}
\def\confYear{2026}

\title{VideoMem:
Enhancing Ultra-Long Video Understanding \\ via Adaptive Memory Management}

\author{
    Hongbo Jin$^{1,3}$ \quad Qingyuan Wang$^1$ \quad Wenhao Zhang$^1$ \quad Yang Liu$^2$ \quad Sijie Cheng$^{2,3}$\thanks{Corresponding Author}\\[1.5ex]
    $^1$School of Electronic and Computer Engineering, Peking University\\
    $^2$Department of Computer Science and Technology, Tsinghua University \\
    $^3$RayNeo.AI 
}

\begin{document}
\maketitle
\begin{abstract}
Ultra-long video understanding remains an open challenge, as existing vision language models (VLMs) falter on such content due to limited context length and inefficient long-term memory retention.
To address this, recent works have attempted to construct external knowledge bases and corresponding retrieval agumented generation (RAG) systems, yet these incur enormous storage and computational overhead.
In this paper,
we propose \textbf{VideoMem},
a novel framework that pioneers models long video understanding as a sequential generation task via adaptive memory management.
Specifically, VideoMem dynamically updates a global memory buffer, which adaptively retains critical information while discarding redundant content across the video timeline.
To efficiently train VLMs for such long-term tasks, VideoMem integrates the Progressive Grouped Relative Policy Optimization (PRPO) algorithm, equipped with two core modules:
Progressive State Propagation (PSP) adaptively retains valid current states, propagates them to the next rollout step, and gradually narrows the model’s exploration space.
Temporal Cascading Reward (TCR) further alleviates reward sparsity, improving sample utilization and accelerating convergence.
Extensive experiments demonstrate that VideoMem significantly outperforms existing open-source models across diverse benchmarks for ultra-long video understanding tasks.
\end{abstract}    
\section{Introduction}
\label{sec:intro}
Nowadays, Vision-Language Models (VLMs;~\citep{liu2023visual,li2023blip,alayrac2022flamingo,li2024llava}) have attained remarkable advancements across a wide range of image-centric downstream tasks~\citep{bai2025qwen2,wang2025internvl3}.
However, processing video content, particularly ultra-long videos, remains a prominent challenge for their core architectures~\citep{wang2025lvbench,fu2025video}.
This phenomenon stems from two fundamental bottlenecks:
(i) \textbf{limited context length}~\citep{chenlongvila,liu2024kangaroo},
restricted by the quadratic computational cost of attention mechanisms;
(ii) \textbf{inefficient memory modeling}~\citep{he2024ma,song2024moviechat,wang2025videollamb},
which results in catastrophic forgetting of early video content.
These constraints impede the development of critical applications such as long-form video summarization and question-answering over extended timelines, both of which demand comprehensive and holistic video comprehension.

Existing long-video understanding approaches can be broadly categorized into three paradigms.
Compression-based methods reduce input dimensionality through frame sampling or feature compression~\cite{jin2024chat, li2024llama,shenlongvu,wang2025dynamic,wu2025marc},
yet inevitably discard fine-grained temporal details.
Static memory mechanisms employ pre-defined rules to manage long-term memory~\cite{song2024moviechat,zhang2024flash,wang2025episodic,he2024ma,qian2024streaming,wang2025videollamb},
but lack adaptability to accommodate diverse video content and task-specfic requirements.
RAG-based methods~\cite{luo2024video,tian2025ego,long2025seeing}
construct external knowledge bases to store full video information and design corresponding retrieval-augmented generation (RAG) systems,
which however incur substantial storage and computational overhead.

\begin{figure*}[h]
    \centering
    \includegraphics[width=\linewidth]{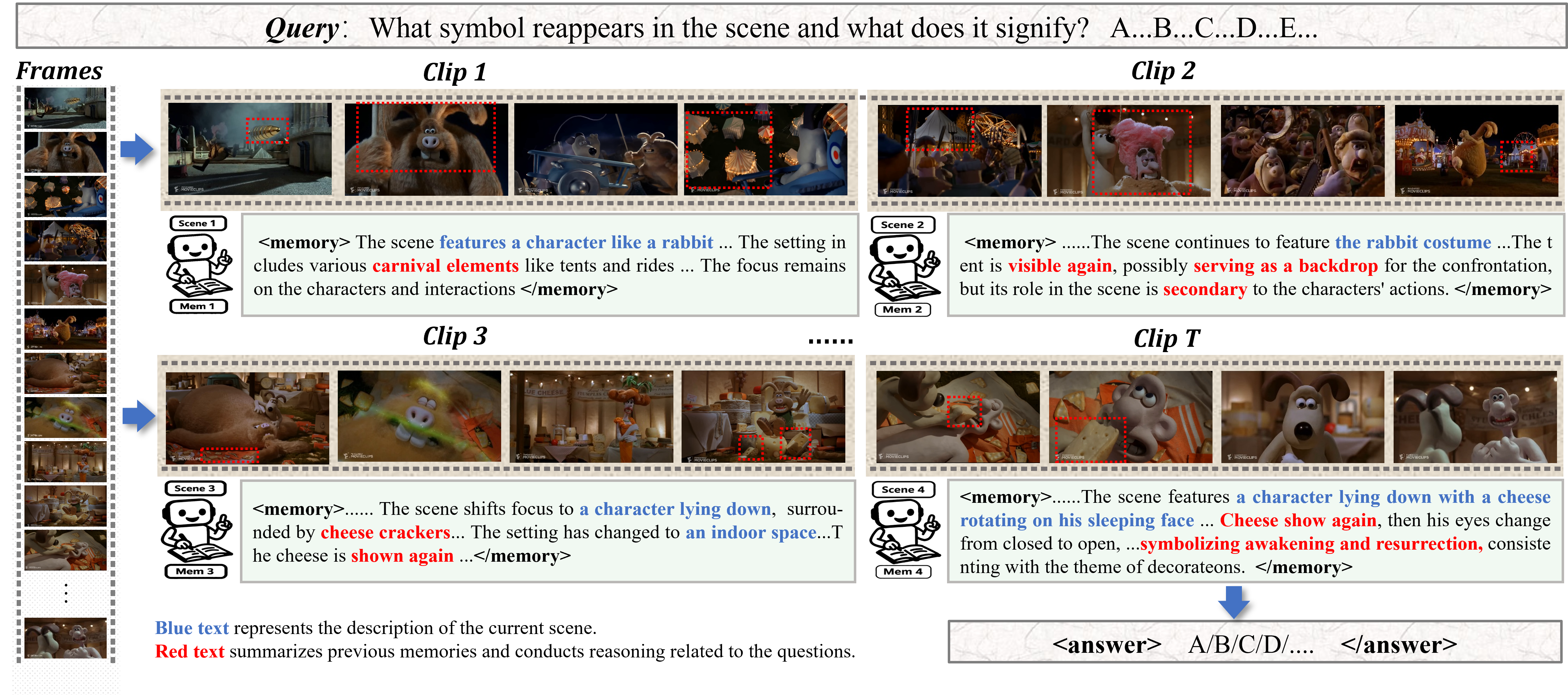}
    \caption{VideoMem demonstrates amazing ability to capture key information and perform long-term memory.}
    \label{fig:sequential}
\end{figure*}% % % % % % % % 

To address these challenges, we propose \VideoMem, a framework for ultra-long video understanding via adaptive memory management.
This core component dynamically retains critical video details to mitigate forgetting while discarding redundant content to control resource consumption.
In Figure~\ref{fig:sequential}, \VideoMem pioneers framing long-video understanding as a sequential generation task conditioned on the query.
Specifically, the long video is first split into segments and are then processed iteratively:
at each step, the model dynamically updates a global memory buffer using the current segment, while reserving key temporal details from prior segments.
After processing the entire long-video, the model synthesizes all accumulated memories to generate the final answer.

To train this framework, reinforcement learning (RL) is a natural fit for such sequential tasks.
However, directly applying prior methods, like GRPO~\cite{shao2024deepseekmath}, to ultra-long videos processing is hindered by sparse and delayed reward signals, as well as prohibitive training computational costs.
We thus propose Progressive Grouped Relative Policy Optimization (PRPO), an algorithm that enable efficient and effective RL training for this setting.
In detail, PRPO reduces training overhead by progressively propagating valid states and narrowing the exploration space, namely progressive state propagation (PSP).
Furthermore, it incorporates a temporal cascading rewards (TCR) mechanism to provides denser, stage-aware feedback, effectively mitigating the sparse reward problem inherent to long-term tasks.

To extensively validate our proposed framework, we conduct experiments on a comprehensive suite of standard and challenging long-video benchmarks, including VideoMME~\citep{fu2025video}, LongVideoBench~\citep{wu2024longvideobench}, MLVU~\citep{zhou2024mlvu}, LVBench~\citep{wang2025lvbench}, LongTimeScope~\citep{zohar2025apollo2}.
On one hand, leveraging adaptive memory management, \VideoMem achieves state-of-the-art (SOTA) performance among 17 comparable open-source models across all long-video benchmark.
On the other hand, PRPO excels in training efficiency: it outperforms GRPO by 3.1× in computational efficiency, cuts convergence steps by 30\% (i.e., 0.7× the original), and yields notably improved training stability.

% In summary, our contributions are threefold:
% \begin{itemize}
%     \item We propose a novel framework VideoMem that enables adaptive memory management for ultra-long video understanding pioneeringly.
%     \item We propose a novel RL algorithm PRPO that enhances training efficiency and stability for long-term tasks through a combination of temporal cascading rewards and progressive state propagation.
%     \item We demonstrate through extensive experiments that VideoMem, trained with PRPO, achieves superior performance on several challenging long-video benchmarks, validating the effectiveness of our approach.
% \end{itemize}

\section{Related Work}
\label{sec:relat}

\subsection{Long Video Understanding}
Long video understanding poses challenges that stem from highly redundant visual information and inefficient temporal dependency modeling.
Expanding context length is an intuitive solution,
which directly enlarges model’s temporal receptive field.
Methods such as Gemini\cite{team2023gemini},
Kangaroo\cite{liu2024kangaroo},
and LongVILA \cite{chenlongvila} achieve this via efficient positional encoding or sliding-window.
Token compression approaches such as VideoAgent \cite{fan2024videoagent},
LongVU \cite{shenlongvu},
LLaVA-onevision \cite{li2024llava}, PLLaVA \cite{xu2024pllava} and VideoXL-2 \cite{qin2025video} attempt to reduce the sequence length via pooling mechanism or similarity computation,
while tend to lose fine-grained details.
Recent studies focus on building RAG systems for long videos.
Ego-R1 \cite{tian2025ego} constructs a knowledge base in a multi-level hierarchy,
and trains VLMs to retrieve from it.
M3-Agent \cite{long2025seeing} trains two models separately:
one learns to store while the other learns to retrieve.
Different from the above methods,
we enhance long video understanding via adaptive memory management without external databases or handcrafted selection rules.

\subsection{Reinforcement Learning}
RL has become a cornerstone for LLMs to align with human preferences and enhance reasoning capabilities \cite{ouyang2022training}.
PPO \cite{schulman2017proximal} optimizes policies through actor-critic frameworks,
balancing exploration and exploitation to improve generation quality.
Recent GRPO \cite{shao2024deepseekmath} further reduces computation overhead by eliminating critic model and estimating advantages relative to grouped samples.
In video domain,
Video-R1 \cite{feng2025video} explores the R1 paradigm \cite{guo2025deepseek} to incentivize video reasoning.
LongVILA-R1 \cite{chen2025scaling} introduces a full-stack framework that scales up reasoning in long videos.
TinyLLaVA-Video-R1 \cite{zhang2025tinyllavavideor1smallerlmmsvideo} explored the reasoning potential of small-scale VLMs.
However,
standard RL approaches struggle with exponential exploration space and reward sparsity in ultra-long video sequences.
Our work incorporates temporal cascading rewards and progressive state propagation specifically designed for such challenges.

\section{VideoMem}
\label{sec:method}

\begin{figure}[t!]
    \centering
    \includegraphics[width=\linewidth]{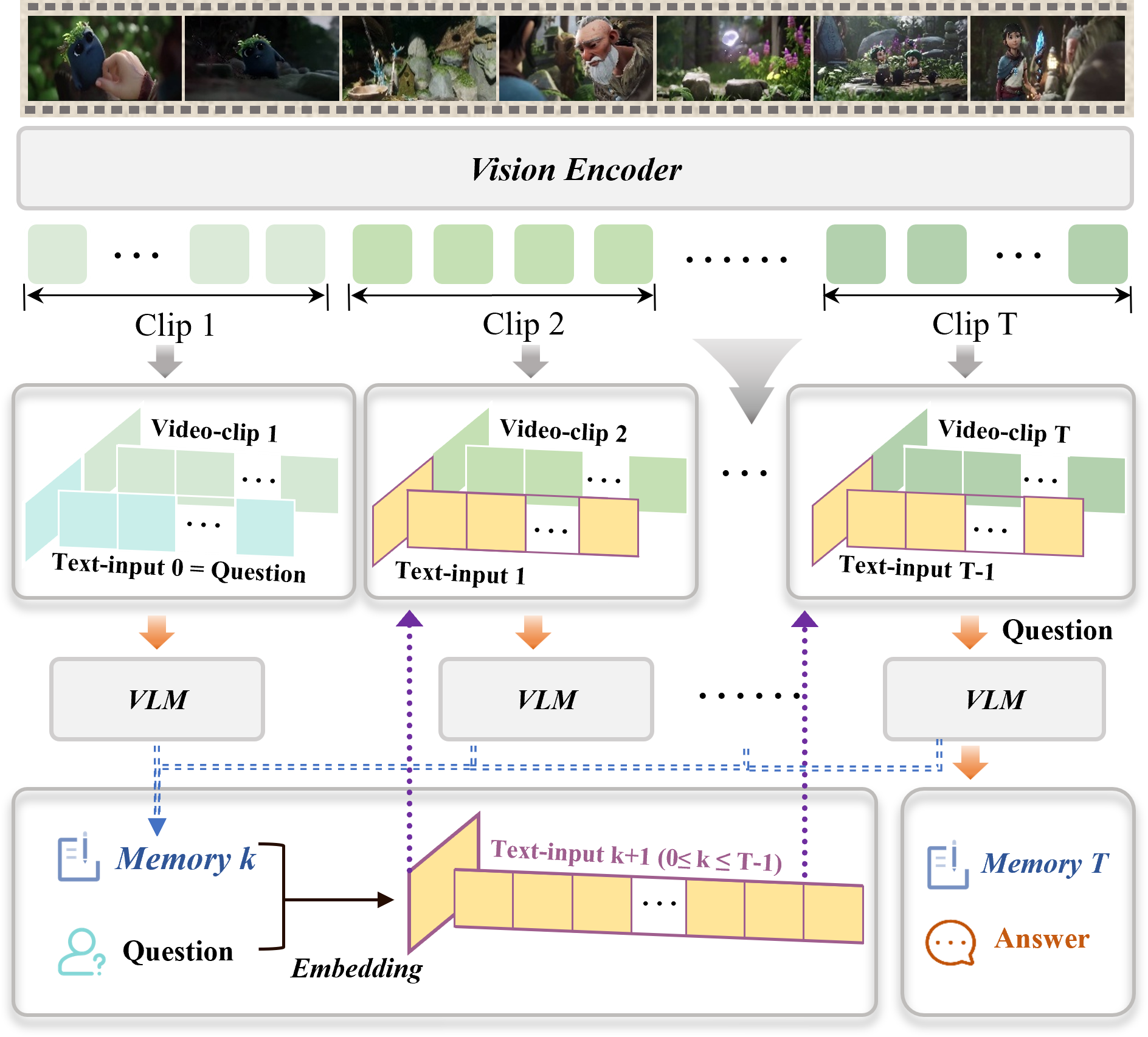}
    \caption{An overview of VideoMem framework.
    The video are divided into multiple clips and fed into the model iteratively.
    The question, video clips are concatenated with the memory of each turn's model output as input for the next turn.
    }
    \label{fig: overall pipeline}
\end{figure}

\begin{figure*}[t!]
    \centering
    \includegraphics[width=0.95\textwidth]{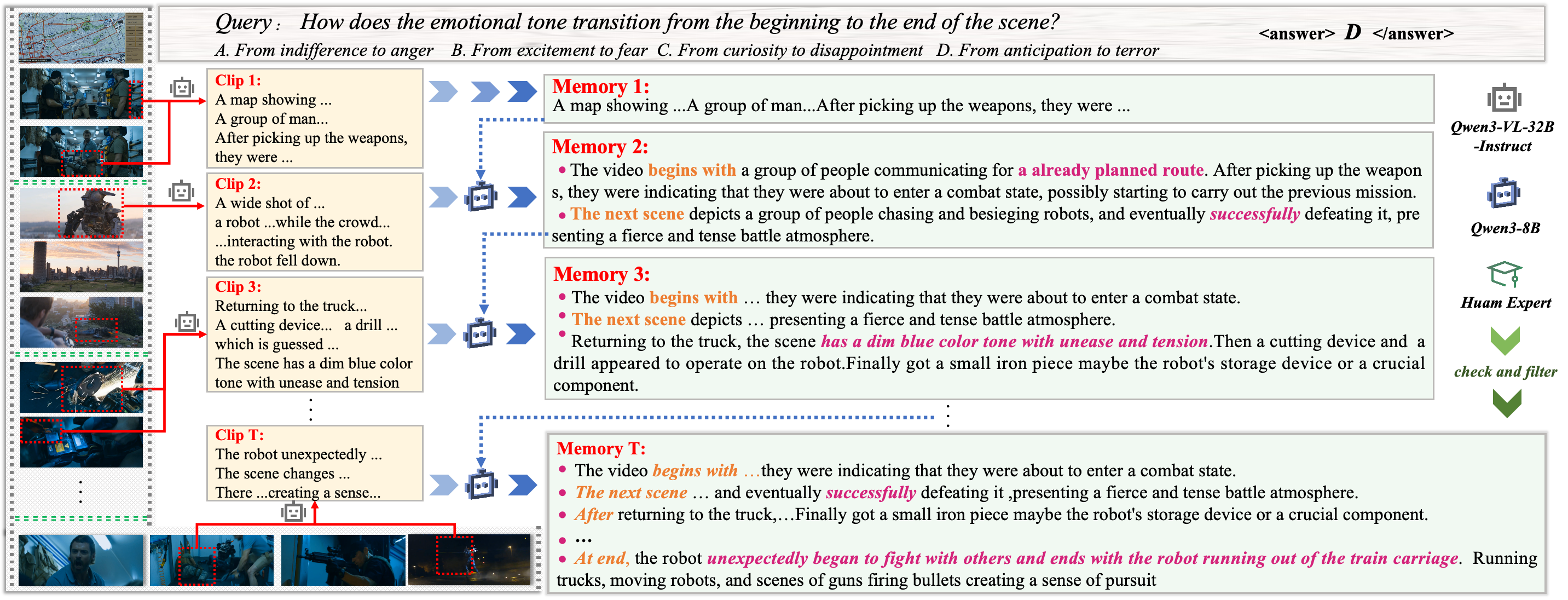}
    \caption{Chain-of-memory data construction pipeline.
    By combining multiple models,
    multi-stage generation,
    and human supervision,
    we ensure high-quality of video chain-of-memory pairs.}
    \label{fig: data pipeline}
\end{figure*}

In this section, we elaborate on our proposed framework \VideoMem.
First, we formulate ultra-long video understanding task as a sequential generation task and outline adaptive memory management,
as shown in Figure \ref{fig: overall pipeline}.
Second, we design an automated pipeline to construct chain-of-memory data and perform cold start training for enhanced stability, as illustrated in Figure~\ref{fig: data pipeline}.
Finally, we propose a RL-based training algorithm, Progressive Grouped Relative Policy Optimization (PRPO), which comprises two core components: temporal cascading reward (TCR) and progressive state propagation (PSP).

\begin{algorithm}[h]
\caption{Progressive Grouped Relative Policy Optimization (PRPO)}
\label{alg:prpo}
\textbf{Require:} Policy $\pi_\theta$, video segments $V = \{V_1,\dots,V_T\}$, question $Q$, ground-truth answer $\text{Ans}$, Group count $G$, token budget $L_{\max}$, coefficients $\alpha$, $\beta$ \\
\textbf{Ensure:} Updated policy $\pi_\theta$
\begin{algorithmic}[1]
\STATE Initialize memory $M \leftarrow {}$\texttt{""}
\STATE Initialize trajectory buffer $\mathcal{B} \leftarrow \emptyset$
\FOR{each example in dataset}
    \STATE $Q, V, \text{Ans} \leftarrow \text{example}$
    \FOR{$t = 1$ \textbf{to} $T$}
        \STATE $\{O_i\}_{i=1}^G \leftarrow \text{Rollout}(\pi_\theta, Q, V_t, M)$ \COMMENT{Generate $G$ trajectories}
        \STATE $\{\hat{a}_i\}_{i=1}^G \leftarrow \text{ExtractAnswer}(Q, \{O_i\}_{i=1}^G)$ \COMMENT{Provisional answer from $<$\texttt{answer}$>$}
        \STATE $\text{ConsReward}_i \leftarrow \mathbb{I}[\hat{a}_i = \text{Ans}]$ \COMMENT{$1$ if matches $\text{Ans}$, $0$ otherwise}
        \STATE $\text{FmtReward}_i \leftarrow \mathbb{I}[\text{CheckFormat}(O_i)]$ \COMMENT{$1$ if $<$\texttt{memory}$>\dots<$/\texttt{memory}$>$ $<$ \texttt{answer}$>\dots<$/\texttt{answer}$>$, $0$ else}
        \STATE $\text{MemPenalty}_i \leftarrow \max(0, \| \text{ExtractMemory}(O_i) \|_{\text{tokens}} - L_{\max})$
        \STATE $R_i \leftarrow \alpha \cdot \text{ConsReward}_i + \text{FmtReward}_i - \beta \cdot \text{MemPenalty}_i$
        \STATE $\{A_i\}_{i=1}^G \leftarrow \text{ADV}(\{R_i\}_{i=1}^G)$ \COMMENT{Group-relative advantage}
        \STATE $\mathcal{B} \leftarrow \mathcal{B} \cup \{ (O_i, A_i) \}_{i=1}^G$ \COMMENT {Reserved for policy updates}
        \IF{$t < T$}
            \STATE $M \leftarrow \text{ExtractBestMemory}(\{O_i\}_{i=1}^G, \{A_i\}_{i=1}^G)$ \COMMENT{Sample $m_j \sim p_j \propto A_j^{\text{rel}}$}
        \ENDIF
    \ENDFOR
    \STATE Update $\pi_\theta$ via GRPO-style objective using $\mathcal{B}$
\ENDFOR
\end{algorithmic}
\end{algorithm}

\subsection{Adaptive Memory Management}
Ultra-long video understanding grapples with two inherently conflicting bottlenecks: redundant temporal information in ultra-long content demands compression to save computation yet risks losing critical details, while inefficient long-term memory modeling requires continuously retaining key information for holistic comprehension yet exacerbates redundancy and computational burden.
To address these issues, we frame ultra-long video understanding as a query-conditioned sequential generation task augmented with adaptive memory management.
This method dynamically retains key information, discards redundancy, and accumulates temporal context across video segments, enabling the model to holistically comprehend ultra-long videos without catastrophic forgetting or prohibitive costs.

In Figure~\ref{fig: overall pipeline}, we first split a given ultra-long video into a sequence of equally long clips $\{V_1,V_2,...,V_T\}$. 
We then iteratively feed these clips to the VLM one on at a time, processing clip $V_t$ at step $t$.
At each step $t$, the VLM takes three types of inputs: the current video clip \(V_t\), the user’s query $Q$, and a global memory buffer \(M_{t-1}\).
Given these inputs and the prompt detailed in Appendix~\ref{sec:prompt template}, the VLM retains critical temporal details (e.g., recurring symbols, plot transitions) and prunes redundant content (e.g., static backgrounds) to form the updated global memory buffer \(M_{t}\).
This adaptive memory management balances current information with prior context, resolving the ``forgetting vs. redundancy'' dilemma.
Finally, the global memory buffer \(M_T\) adaptively accumulates key information from all clips \(\{V_1, ..., V_{T}\}\), enhancing the accuracy of the final answer.
% The model are optimized to better update memories, preserve effective information in memory, and analyze memories to arrive at the correct answers,
% as shown in Figure \ref{fig: overall pipeline}.

This sequential generation task can be naturally formulated as a reinforcement learning (RL) problem, where the policy $\pi_\theta$ is optimized to generate high-quality trajectories.
However, challenges arise in two key aspects:
first, the ultra-long video setting exacerbates inherent reward sparsity and demands improved training efficiency;
second, the model requires robust foundational memory capabilities to avoid unproductive exploration during RL training.
Therefore, we detail two tailored training components in the following subsections: a cold-start stage and the Progressive Grouped Relative Policy Optimization algorithm.

\begin{figure*}[h]
    \centering
    \includegraphics[width=0.9\textwidth]{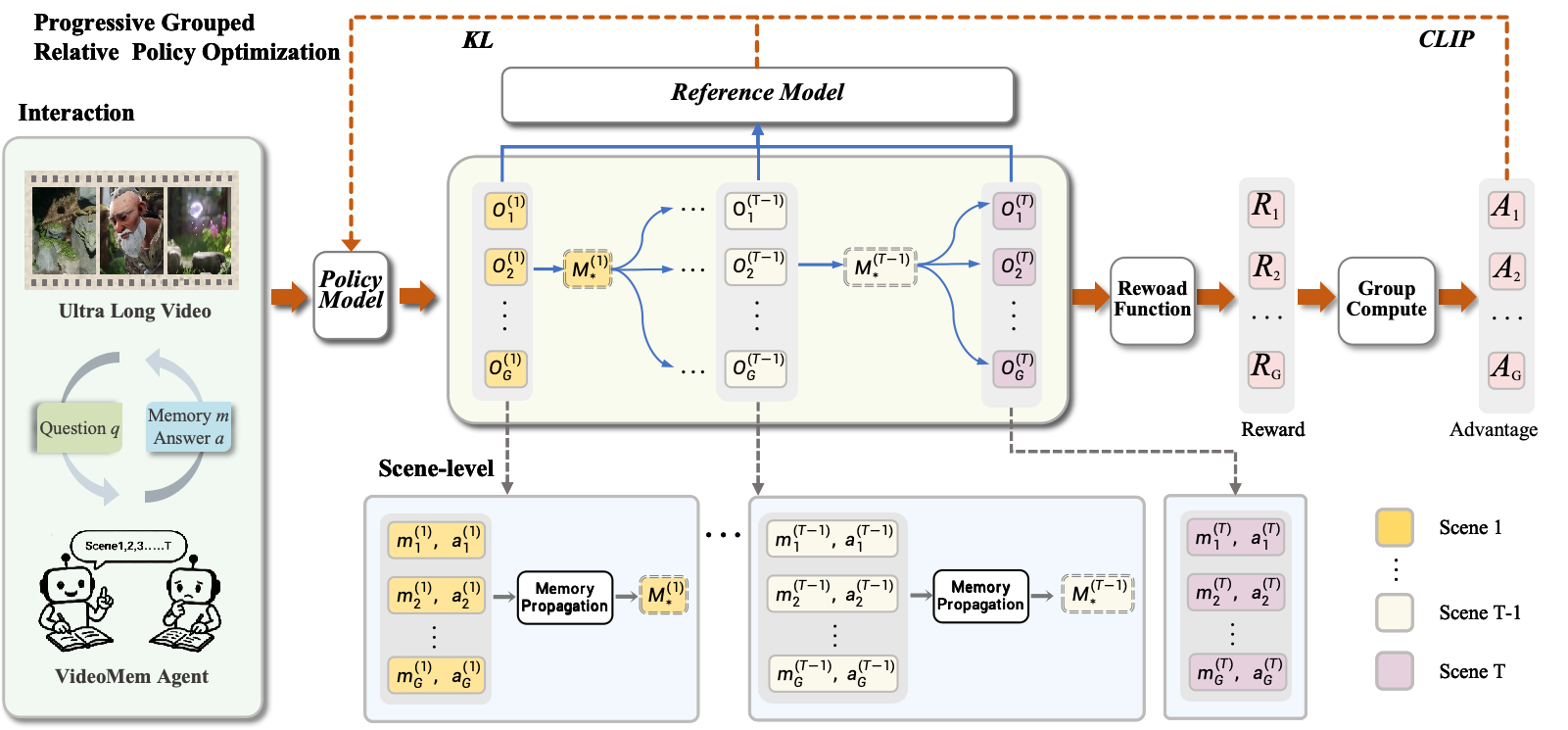}
    \caption{rollout and policy optimization pipeline of PRPO}
    \label{fig: prpo pipeline}
\end{figure*}

\subsection{Cold Start Stage}
Following Deepseek-R1~\citep{shao2024deepseekmath}, we introduce a cold start stage prior to RL to equip the base model with fundamental capabilities for compressing ultra-long videos and retaining their key information. 
By pre-aligning the model with basic memory management heuristics, it further accelerates the RL process, enabling more efficient discovery of high-performance policies—thereby stabilizing and enhancing the subsequent training pipeline.

To automatically construct our chain-of-memory data, we first sample video–question–answer pairs from VideoMarathon \cite{lin2025unleashinghourscalevideotraining} dataset.
VideoMarathon is a comprehensive open-source long video training dataset, covering a wide range of video distributions with 1,573k multiple-choice questions and 1,726k open-ended questions.
Each video is split into multiple equal long clips, which are sequentially fed into \texttt{Qwen3-VL-32B-Instruct} to generate semantic summaries.
These summaries capture core narrative or factual content according to the context of specific video clips.
Then, the clip-wise summaries are fed to \texttt{Qwen3-8B} \cite{yang2025qwen3} iteratively
to filter out irrelevant information and generate coherent memories with manual checking.
This yields a high-quality chain-of-memory dataset.
Prompt templates are illurated in Appendix~\ref{sec:prompt template}.

Finally, we fine-tune the base model on these high-quality chain-of-memory instances using standard supervised instruction tuning to initialize its memory management ability.
This cold start phase equips the model with the fundamental capability to construct coherent and concise memory representations.

\subsection{Progressive Grouped Relative Policy Optimization}

While Grouped Relative Policy Optimization (GRPO;~\citep{shao2024deepseekmath}) lays a solid foundation for policy optimization without step-wise reward modeling,
it encounters notable challenges when adapted to long-term tasks.
The primary issues are the exponential expansion of the exploration space and exacerbated sparse reward signals inherent to extended temporal sequences.
To mitigate these limitations, as illustrated in Figure~\ref{fig: prpo pipeline}, we propose Progressive Grouped Relative Policy Optimization (PRPO).
This algorithm is integrated with two key innovations:
temporal cascading reward (TCR) and progressive state propagation (PSP).

\subsubsection{Temporal Cascading Reward}

A major challenge in long-term RL is the sparsity and delay of reward signals:
A reward for correctly answering a question grounded in early video content may only be observed after processing the entire sequence,
rendering credit assignment highly non-trival.
To address this, the temporal cascading reward (TCR) mechanism decomposes sparse final rewards into denser intermediate rewards at each step, delivering more frequent feedback to guide memory management.
Specifically, at each step $t$, the model generates a provisional answer $ \hat{a}_t $ conditioned on the current video clip $V_t$, memory buffer $ M_{t-1} $ and query $ Q $. 
The immediate reward $ \hat{R}_t $ for each trajectory is then computed as the sum of three components:
\begin{equation}
    \hat{R}_t = \alpha \cdot R_{\text{cons}} + R_{\text{format}} - \beta \cdot \text{MemPenalty}
    \label{eq: reward}
\end{equation}

\begin{itemize}
    \item 
    $R_{\text{cons}} = \mathbb{I}[\hat{a}_t = a]$:
    For multiple-choice QA tasks, consistency reward is 1 if $ \hat{a}_t $ exactly matches the ground-truth answer $a$ and 0 otherwise.
    \item $ R_{\text{format}} $:
    A binary reward, where the reward is equal to 1 if the output conforms to the required format $\langle \texttt{memory} \rangle \cdots \langle \texttt{memory} \rangle \langle \texttt{answer} \rangle \cdots \langle \texttt{answer} \rangle$, 0 otherwise.
    \item $\text{MemPenalty}=\max(0, \| M_t \| - L_{\max})$:
    Penalizes the length of memory buffer exceeding threshold $ L_{\max} $ to prevent bloat while preserving efficiency.
\end{itemize}
Hyperparameters $ \alpha $ and $ \beta $ control the relative importance of consistency and memory efficiency, respectively.
These rewards are used to compute group-relative advantages $ \hat{A}^{\text{rel}}_t $ following the same procedure as GRPO for policy update,
detailed in Appendix~\ref{sec:group computation}.
This TCR mechanism can alleviate reward sparsity, improve sample utilization, and accelerate convergence.

\subsubsection{Progressive State Propagation}
Moreover, another key challenge in long-term RL is the exponential expansion of the exploration space as the sequence accumulates.
Directly applying GRPO would incur prohibitive computational costs, especially for ultra-long video sequences.
To mitigate this cost barrier, the Progressive State Propagation (PSP) mechanism acts as a memory-aware trajectory selection strategy that adaptively narrows the exploration space while retaining high-reward paths.

Let $ M_t = \{m_1, m_2, \dots, m_G\} $ denotes the set of global memory buffers extracted from the $ G $ trajectories at step $ t $.
Instead of retaining all $ G $ states,
PRPO selects a single memory buffer $ M_t^* $ to propagate to the next turn generation,
conditioned on the group-relative advantage $ \hat{A}_i^{\text{rel}} $ of each trajectory:

\begin{equation}
    M_t^* = \text{Sample}(M_t, \, p_j \propto \hat{A}_j^{\text{rel}})
\end{equation}
The sampling probability is softened with a temperature $ \tau_t $ that decreases over time:
\begin{equation}
    p_j = \frac{\exp(\hat{A}_j^{\text{rel}} / \tau_t)}{\sum_{k=1}^G \exp(\hat{A}_k^{\text{rel}} / \tau_t)}
\end{equation}
where $ \tau_t = \tau_0 \cdot \gamma^t $ ($ \gamma < 1 $) ensures high exploration early and focused exploitation later.
The selected state $ M_t^* $ is concatenated with the query $ Q $ and the upcoming clip $ V_{t+1} $ to form the input for the next turn rollout.
This progressive narrowing yields two key benefits:
(i) Computational efficiency, as prefilling complexity drops from $ O(N \cdot G) $ to $ O(N) $. Since only one memory path is cached per step,
explained in detail in Section \ref{sec: computation}.
(ii) Training stability, where early noise is filtered out and only trajectories with strong provisional consistency and format compliance propagate, thereby preventing error accumulation.

\section{Experiment}
\label{sec:exp}

\begin{table*}[h]
    \centering
    \caption{Evaluate results on various long video benchmarks,
    including VideoMME, LongVideoBench, MLVU, LVBench, LongTimeScope.
    All benchmarks adopt percentage accuracy as the evaluation metric.
    *Scores from our evaluation using 64 uniform frames,
    to ensure fair comparisons with same context length.
    All other results are from their original paper, with at least 64 frames of input.
    }
    \resizebox{\textwidth}{!}{
    \begin{tabular}{lccccccccc}
        \toprule
        \multirow{2}{*}{\textbf{Models}} & \multirow{2}{*}{\textbf{Size}} &\multicolumn{2}{p{3.2cm}}{\centering \textbf{VideoMME} }& \multirow{2}{*}{ \textbf{LongVideoBench} } & \multirow{2}{*}{ \textbf{MLVU} }  & \multirow{2}{*}{\textbf{LVBench}} & \multirow{2}{*}{\textbf{LongTimeScope}}\\ 
        && Long & Overall \\
        \rowcolor{gray!10} 
        Duration&&30$\sim$60 min & 1$\sim$60 min& 23sec$\sim$60 min & 3$\sim$120 min & 30$\sim$90 min & 300min+  \\
        \midrule
        \rowcolor{gray!10} 
        \multicolumn{8}{c}{\textbf{\textit{Proprietary Models}}} \\
            GPT-4V &-&56.9&60.7&59.1&49.2&-&-\\
            GPT-4o &-&65.3&71.9&66.7&64.6&64.4&-\\
            Gemini-2.5-Pro &-&-&87.0&-&81.2&69.2&-\\
            Seed1.5-VL&8B&-&77.9&74.4&82.1&64.6&- \\
            Eagle2.5&8B &-&72.4&66.4&77.6&-&-\\
        \midrule
        \rowcolor{gray!10} 
        \multicolumn{8}{c}{\textbf{\textit{Open Source Models}}} \\
            Video-LLaVA \cite{lin2024video} &7B&38.1&40.4&39.1&47.3&-&17.6\\
            ShareGPT4Video \cite{chen2024sharegpt4video} &8B&37.9&43.6&39.7&46.4&-&-\\
            LLaVA-Next-Video \cite{liu2024llavanext} &7B&-&46.5&50.5&39.3&-&- \\
            VideoLLaMA2 \cite{cheng2024videollama} &7B&43.8&46.6&-&48.5&-&-\\
            LongVA \cite{zhang2024long} &7B&47.6&54.3&-&56.3&36.2&-\\
            LLaVA-Onevision \cite{li2024llava} &7B& 46.7&58.2&56.4&64.7&26.9&30.2\\
            LLaVA-video \cite{zhang2025llavavideovideoinstructiontuning}&7B&50.6&62.6&58.2&70.8&41.5&34.0\\
            Qwen2.5-VL \cite{bai2025qwen2} &7B&51.6&65.1&54.7&70.2&45.3&40.7 \\
            VideoNSA \cite{song2025videonsa}&7B &-&-&60.0&51.8&-&44.4\\
            Apollo \cite{zohar2025apollo}&7B&-&61.3&58.5&70.9&-&-\\
            NVILA \cite{liu2025nvila} &8B&54.8&64.2&57.7&70.1&44.0&- \\
         Flow4Agent\cite{liu2025flow4agent}&7B&54.2&64.7&60.4&71.4&-&-  \\
            VideoLLaMA3 \cite{zhang2025videollama}&7B&54.1&66.2&59.8&73.0&45.3&39.1 \\
            Qwen2.5-VL&72B&61.2&73.3&-&74.6&47.3&-\\
            InternVL3.5 \cite{wang2025internvl3}&8B&-&66.0&62.1&70.2&-&-\\
            Video-XL2 \cite{qin2025video}&8B &-&66.6&61.0&74.8&48.4&- \\
            % Qwen3-VL&4B&53.8&69.3&&75.3&56.2&27.7\\
            % Qwen3-VL&8B&58.6&71.4&59.1&78.1&58.0&36.2\\
            Qwen3-VL*&8B&58.6&67.9&59.1&71.6&50.6&36.2\\
        \midrule
            \rowcolor[RGB]{207,234,241}
            \textbf{VideoMem*}&8B & \textbf{64.2}&\textbf{73.6}&\textbf{63.3}&\textbf{77.4}&\textbf{58.5}&\textbf{45.1} \\
         \bottomrule
    \end{tabular}
    }
    \label{tab:evaluate results}
\end{table*}

\subsection{Implementation Details}
We choose Qwen3-VL-8B as our base model.
During training,
we divide each video into four equal segments, 
and extract 32 frames from each segment with max pixels not exceeding $128*32*32$.
The number of rollouts $G$ is set to 8.
For temporal cascading reward,
the hyperparameters are set to $\alpha=1.0$,
$\beta=0.005$ and $ L_{\max} =1024$.
We set global batch size to 16 and use the AdamW optimizer.
All experiments are conducted on 16x PPU-ZW810E with 96GB memory.
The complete training process requires 4800 PPU hours or 3200 A100 GPU hours.
During evaluation,
the input frame number is all set to 64,
and the number of video segments in VideoMem is 4.

As for data, we sample 26k records from VideoMarathon dataset,
and label them with chain-of-memory as the cold start training set.
During RL training stage,
we sample 147k multi choice QA pairs from VideoMarathon 
and 49k from LLaVA-Video-178K \cite{zhang2025llavavideovideoinstructiontuning} dataset as training dataset,
emphasis on long videos while also considering short videos.
We adopt this data ratio,
following the best results described in VideoMarathon \cite{lin2025unleashinghourscalevideotraining}.

\subsection{Main Results}
Table \ref{tab:evaluate results} presents our main results across all benchmarks.
VideoMem significantly outperforms various strong baselines,
achieving new state-of-the-art (SOTA) performance among comparable open-source models on diverse long-video benchmarks.
Notably, our model delivers the largest performance gains on the most challenging long-form video benchmarks--LVBench, VideoMME and LongTimeScope--outperforming the base model Qwen3-VL-8B by 7.9\%, 5.7\% and 8.9\%, respectively.
These results are particularly revealing. While the base model exhibits strong static visual comprehension capabilities, it is not inherently optimized for the long-term reasoning required by ultra-long videos.
It processes extended context as a flat sequence, struggling to effectively retain salient information over time. 
In contrast, VideoMem is explicitly trained via PRPO to act as a stateful agent: 
it learns an adaptive memory management policy that decides what critical information from the current clip $V_t$ to integrate into memory $M_t$ and what to discard.
The substantial performance gains provide direct evidence that our adaptive memory mechanism successfully overcomes this core limitation of baseline models.

These findings offer a key insight:
rather than relying solely on expanding context windows, which faces computational scaling challenges, a more efficient and effective paradigm is to reframe long-video understanding as a sequential generation problem.
The fundamental reason for VideoMem's success lies in this shift—from training a model for static, one-shot comprehension to training an agent with a dynamic policy for information compression and propagation. Our work demonstrates that VLMs can be effectively optimized for long-term tasks using tailored RL strategies like PRPO, which successfully manage the sparse-reward and large-exploration-space challenges inherent to this formulation.

\subsection{Core Components Analysis}

We ablate the core components of VideoMem in Table~\ref{tab:ablation}. Removing progressive state propagation (PSP) drops performance by 2.1\%,
confirming its role in stabilizing long-term training.
Removing temporal cascading reward (TCR) causes even larger degradation (4.2-4.9)\%,
leading to severe reward sparsity problem.
And removing the memory penalty will also lead to a certain performance degradation.

\begin{table}[h]
\centering
\caption{Ablation of VideoMem core components.}
\small
% \begin{adjustbox}{width=0.45\textwidth}
\begin{tabular}{l|ccc|cc}
\toprule
\textbf{Method} & \textbf{PSP} & \textbf{TCR} & \textbf{MP} & \textbf{MME}&\textbf{LTS} \\
\midrule
VideoMem & \checkmark & \checkmark & \checkmark & \textbf{73.6} &\textbf{45.1}\\
w/o PSP. & & \checkmark & \checkmark & 71.5& 43.0\\
w/o TCR & \checkmark & & \checkmark & 69.4 &40.3 \\
w/o MP & \checkmark & \checkmark & & 72.0 &43.9\\
SFT  & & & & 67.5&38.2\\
\bottomrule
\end{tabular}
% \end{adjustbox}
\label{tab:ablation}
\end{table}
\subsection{Reward Mechanism Analysis}
Table \ref{tab:temporal reward} further analyzes the impact of temporal cascading reward (TCR) compared to the single terminal reward (TR) used in GRPO.
We observe significantly denser and more stable reward feedback when applying TCR.
PRPO’s temporal reward decomposition increases reward frequency by 4.0× (number of segments was set to 4),
resulting in smoother policy gradients and faster convergence.
We also notice that temporal cascading rewards stabilize training and prevent premature policy collapse.
Empirically,
models trained with TCR converge in about 70\% of the steps required by TR,
highlighting the efficiency of our dense feedback design.

\begin{table}[h]
\centering
\caption{Ablation of temporal cascading reward mechanism.}
\small
\begin{adjustbox}{width=0.48\textwidth}
\begin{tabular}{l|cc|c}
\toprule
\textbf{Reward Type} & \textbf{Reward Density} & \textbf{Convergence Steps} & \textbf{LVBench} \\
\midrule
TR &1.0x  & 1.0x & 54.6 \\
TCR &4.0x & 0.7x & 58.5\\
\bottomrule
\end{tabular}
\end{adjustbox}
\label{tab:temporal reward}
\end{table}

\subsection{Per-Segment Accuracy Analysis}
To verify the performance improvement of VideoMem truly stems from long-term memory,
rather than a “general guess” based on early segments,
we conducted the following experiments on LVBench.
\begin{figure*}[h]
    \centering
    \includegraphics[width=0.95\linewidth]{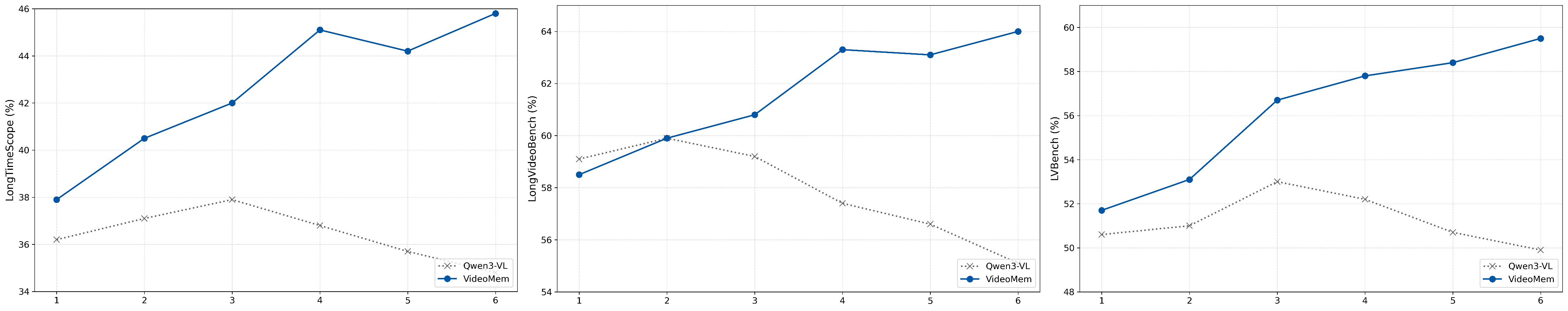}
    \caption{Inference Scalability}
    \label{fig:inference scability}
    
\end{figure*}

We utilize the Temporal Cascaded Reward (TCR) mechanism to generate a provisional answer $\hat{a}_{t}$ at each step $t$,
and record the corresponding "provisional accuracy".
\begin{table}[t]
    \centering
    \caption{Seg1-N represents the accuracy of the model's prediction after N turns of input.}
    \begin{tabular}{ccccc}
        \toprule
            Method & Seg1-1&Seg1-2&Seg1-3&Seg1-4\\
        \midrule
             Qwen3-VL& 50.3 & 48.5 & 46.2& 44.8 \\
             VideoMem & 51.2&55.4 & 57.1&58.5\\
        \bottomrule
    \end{tabular}
    \label{tab:per segment accuracy}
\end{table}
As shown in Table \ref{tab:per segment accuracy},
the accuracy of the baseline model decreased from 50.3\% to 44.8\% with the addition of segments.
This demonstrates catastrophic forgetting, where the model fails to integrate new information to correct early error.
Conversely,
VideoMem's accuracy showed non-trivially increases with each new segment added,
steadily increasing from 51.2\% to 58.5\%.
This strongly demonstrates that VideoMem continuously and proactively utilizes new segments to refine and improve its understanding,
confirming the substantial contribution of later fragments.

\subsection{Cold Start Effiency}

To verify the actual contribution of the cold start phase,
we conducted corresponding ablation experiments in Table \ref{tab:coldstart_ablation}.
We conducted a comparison on three representative long video benchmarks: VideoMME (long subset), LVBench, and LongVideoBench.
All other training hyperparameters (batch size, number of frames, token budget, etc.) are consistent with the main experiment.
Results show that cold start is a necessary training phase that can accelerate model convergence and avoid invalid rollouts.
Detailed diagrams of the training process are shown in the Appendix.

\begin{table}[h]
    \centering
    \caption{Cold Start Efficiency and Key Baseline Comparison.
    We compared five different experimental setups: (1) Qwen3-VL (untrained);
    (2) SFT (fine-tuned on general video instruction data);
    (3) PRPO-only (trained using only PRPO);
    (4) ColdStart-only (SFT training only on our chain-of-memory data);
    (5) VideoMem (trained with both cold start and PRPO).}
    \begin{tabular}{lccc}
        \toprule
        Method & VideoMME & LVBench & LVB  \\
        \midrule
        Qwen3-VL &58.6&50.6&59.1\\
        SFT & 58.1  & 51.7 &58.5 \\
        PRPO-only & 59.8 & 53.4 & 59.5\\
        ColdStart-only & 61.9  & 54.4 &60.7  \\
        VideoMem & \textbf{64.2} & \textbf{58.5}  &\textbf{63.3}\\
        \bottomrule
    \end{tabular}
    \label{tab:coldstart_ablation}
\end{table}

\subsection{Inference Scalability}
While the number of video segments during training is fixed to $T=4$,
to ensure stable RL convergence and balanced reward density,
the learned memory policy generalizes seamlessly to arbitrary segment counts during inference. 

To verify this scalability,
we conducted analysis on LongTimeScope, LongVideoBench and LVBench,
shown in Figure \ref{fig:inference scability}.
Fix the number of segments to 4 during training while evaluate its performance during inference with different numbers of segments $T = {1,2,3,4,5,6}$.
Simultaneously, we compared the performance of the baseline model (Qwen3-VL) under the same settings.

The performance of baseline (Qwen3-VL) quickly reaches a bottleneck around $T=3$,
and shows a rapid decline,
indicating that it cannot effectively utilize long term memory context.
In contrast,
VideoMem exhibits clear positive correlation scalability:
as the number of segments increases,
VideoMem is able to integrate more contextual information,
and its accuracy steadily improves accordingly.

\subsection{Computational Efficiency}
\label{sec: computation}
VideoMem achieves a 3.1x speedup in training compared to vanilla GRPO while maintaining superior performance,
shown in Table \ref{tab:computation efficiency}.
The key lies in \textbf{PSP}:
at each turn,
a memory state is selected based on relative strengths to serve as the starting state for next turn,
avoiding the expensive "prefill" operation for all $G$ paths. This reduces the number of pre-filling calls from $O(N*G)$ to $O(N)$
(only once per segment).
While decoding
(i.e., token-level generation for each candidate trajectory)
still requires $O(N*G)$ times,
its computational overhead is far less than pre-filling.
The evaluation results did not decrease but even improved,
making our approach practical for large-scale deployment.

\begin{table}[h]
    \centering
    \small
    \caption{Computational Efficiency}
    \begin{tabular}{c|ccc}
        \toprule
            Method & Train Speed $\uparrow$ & VideoMME &LVBench\\
        \midrule
             Qwen3-VL& - & 67.9 & 50.6\\
             GRPO & 1.0x &72.0& 56.7\\
             PRPO & 3.1x&73.6& 58.5\\
        \bottomrule
    \end{tabular}
    \label{tab:computation efficiency}
\end{table}

\section{Discussion}
\textbf{Generalizability of VideoMem to Open-ended Tasks.}
A key avenue for future work is to generalize our framework to open-ended QA tasks.
The core sequential processing and adaptive memory management of VideoMem  are task-agnostic.
We hypothesize that by replacing the binary reward $R_{cons}$ within TCR with a score from an LLM-as-a-judge,
PRPO can effectively train VLMs to generate high-quality,
free-form answers,
confirming that VideoMem learns a deep, compressible understanding of video content.

\noindent\textbf{Generalizability of PRPO to Other Long-term Tasks.}
PRPO algorithm is a general-purpose optimization strategy for sequential decision-making under sparse rewards.
Its core components,
PSP and TCR,
address fundamental challenges in long-term tasks, independent of input modality.
We believe PRPO holds significant promise for other domains,
such as long-document QA,
where text chunks replace video clips $V_t$.
Furthermore,
PRPO could be applied to multi-step agents,
where PSP prunes the exponential exploration space of reasoning paths,
and TCR provides denser feedback on intermediate sub-goals, effectively extending its application far beyond video understanding.

\section{Conclusion}
\label{sec:conc}

In this paper,
we introduced VideoMem,
a novel framework designed to address the significant challenges of ultra-long video understanding.
We pioneeringly model this problem as a sequential generation task,
and train VLMs to progressively compress video information and maintain effective long-term memory.
We also propose Progressive Grouped Relative Policy Optimization (PRPO) algorithm,
a novel reinforcement learning strategy tailored for this complex task.
PRPO enhances training efficiency and stability through two key innovations:
Progressive State Propagation (PSP),
which adaptively narrows the model's exploration space to reduce computational costs, and Temporal Cascading Reward (TCR),
which provides denser,
intermediate feedback to effectively alleviate the reward sparsity problem.
Experiments demonstrate that VideoMem achieves new state-of-the-art performance among comparable open-source models on a wide range of challenging long-video benchmarks,
showing particularly strong gains on hour-long tasks.
Results validate that our approach offers a promising solution for VLMs that can genuinely comprehend and reason over ultra-long video content.
Hope that our approach can provide valuable insights
{
    \small
    \bibliographystyle{ieeenat_fullname}
    \bibliography{main}
}

% WARNING: do not forget to delete the supplementary pages from your submission 
% \input{sec/X_suppl}

\end{document}